\documentclass[10pt,twocolumn,letterpaper]{article}

\usepackage{iccv}
\usepackage{times}
\usepackage{epsfig}
\usepackage{graphicx}
\usepackage{amsmath}
\usepackage{amssymb}

\usepackage{array}
\usepackage{caption}
\usepackage{booktabs} 
\usepackage{subcaption}
\usepackage{mathrsfs}
\usepackage{threeparttable}

\usepackage{epsfig}
\usepackage{graphicx}
\usepackage{algorithm}
\usepackage{algorithmic}
\usepackage{amssymb}
\usepackage{mathrsfs} 
\usepackage{amsmath,bm}
\usepackage{array,multirow}
\usepackage{mdwlist}
\usepackage{paralist}
\usepackage{url}
\usepackage{color}
\usepackage{caption}
\usepackage{setspace}
\usepackage{textcomp}
\usepackage{bbding}
\usepackage{booktabs}
\usepackage{threeparttable}

\usepackage[pagebackref=true,breaklinks=true,letterpaper=true,colorlinks,bookmarks=false]{hyperref}

\iccvfinalcopy 

\def\iccvPaperID{xxx} 

\ificcvfinal\fi
\begin{document}

\title{Cross-dataset Training for Class Increasing Object Detection}

\author{Yongqiang Yao, Yan Wang, Yu Guo, Jiaojiao Lin, Hongwei Qin, Junjie Yan\\
  SenseTime \quad \\
{\tt\small \{yaoyongqiang,wanyan1,guoyu,linjiaojiao,qinhongwei,yanjunjie\}@sensetime.com}}


\maketitle

\begin{abstract}
  We present a conceptually simple, flexible and general framework for cross-dataset training in object detection. Given two or more already labeled datasets that target for different object classes, cross-dataset training aims to detect the union of the different classes, so that we do not have to label all the classes for all the datasets. By cross-dataset training, existing datasets can be utilized to detect the merged object classes with a single model. Further more, in industrial
    applications, the object classes usually increase on demand. So when adding new classes, it is quite time-consuming if we label the new classes on all the existing datasets. While using cross-dataset training, we only need to label the new classes on the new dataset. We experiment on PASCAL VOC, COCO, WIDER FACE and WIDER Pedestrian with both solo and cross-dataset settings. Results show that our cross-dataset pipeline can achieve similar impressive performance simultaneously on these datasets compared with training independently.
\end{abstract}

\section{Introduction}

Object detection is one of the most important computer vision tasks. Essentially, 
object detection is a joint task of classification and localization. For a long history,
the community spend a lot of efforts solving the problem of face detection. In the last decade, however, 
general object detection becomes an very active research topic. 

This phenomenon results from three aspects: better representations, better detection frameworks and better supervision.
The best bounding box mAP on COCO~\cite{lin2014microsoft} in the year of 2012 was 5~\cite{ross-slides-eccv2018}, while the mAP rises to 51.3 in 2018~\cite{ross-slides-eccv2018, he2017mask, cai2017cascade}.
The most rapid progress comes from better representations, especially in the deep learning era, from AlexNet~\cite{krizhevsky2012imagenet} all the way to ResNet~\cite{he2016deep} and its variations.
Detection frameworks also have made great progress, from Viola Jones~\cite{viola2001rapid}, DPM~\cite{felzenszwalb2010object} to the modern R-CNN series~\cite{girshick2014rich, girshick2015fast, ren2015faster, he2017mask, lin2017feature}. 
Better supervision results from the dataset blossom, for example, Pascal VOC~\cite{everingham2010pascal}, COCO~\cite{lin2014microsoft}, WIDER FACE~\cite{yang2016wider}, ImageNet~\cite{deng2009imagenet}, Visual Genome~\cite{krishna2017visual} and Open Image Dataset~\cite{kuznetsova2018open}.

\begin{figure}
    \centering 
    \includegraphics[width=0.49\textwidth]{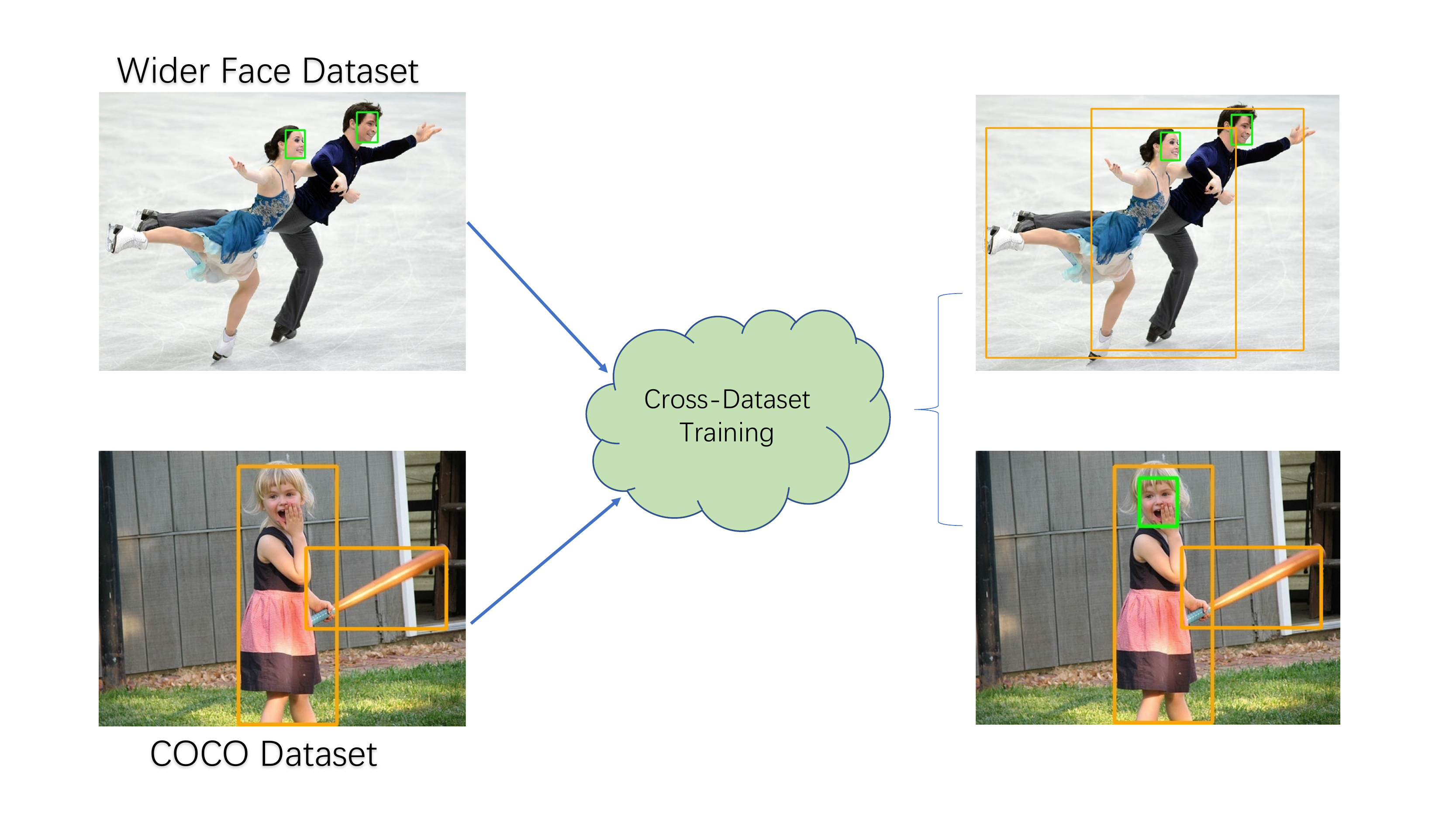}
    \caption{Motivation of cross-dataset training. A single model is trained across multiple existing datasets with different object classes and can be used to detect the union of the object classes from all the datasets, which saves the heavy burden of labelling new classes on all the existing datasets.} 
    \label{fig:intro}
\end{figure}

The trend is that more and more object classes are in demand, as well as the data amount. In practice, new object classes need to be added continuously. 
But it costs time and human resources if we label the new classes for all the existing data. It would be much flexible if we can continuously add new data labeled with new classes. Eventually, a life long learning system can utilize the incomplete labeled datasets. In this paper, we propose to solve this problem with a novel concept called cross-dataset training.

Cross-dataset training aims to utilize two or more datasets labeled with different object classes to train a single model that can performs well on all the classes, as shown in Figure~\ref{fig:intro} . Generally, a face detection model is trained on WIDER FACE to perform face detection only, and a general object detection model is trained on COCO to perform 80 class object detection. By cross-dataset training on WIDER FACE and COCO, our goal is a single model that has the same backbone and can detect 81 classes without accuracy loss.

For cross-dataset object detection, simply concatenating the labels is unreasonable. The first reason is that labels may be duplicated, making it necessary to first merge
the identical labels across datasets. The second reason is the possible conflicts among the positive and negative samples from different datasets. For example, the negative samples from a face detection dataset may contain a large amount of human body, which makes the task very confusing if cross-trained with a human detection dataset.
Considering these two aspects, we propose a novel cross-dataset training scheme specially designed for object detection, which is composed of four steps: 

1) merge duplicated labels across datasets; 

2) generate a hybrid dataset through label concatenation but still keep the original partition information of every image; 

3) build an avoidance relationship across partitions such as face-negative versus human-positive;

4) train the detector with this hybrid dataset where the loss is calculated according to this avoidance relationship.

We experiment on several popular object detection datasets to evaluate cross-dataset training. First we choose WIDER FACE~\cite{yang2016wider} and WIDER Pedestrian~\cite{wider-pedestrian}. We train a baseline model for face detection on WIDER FACE and another for pedestrian detection on WIDER Pedestrian, respectively. When performing cross-dataset training using our training scheme, we get a single model that simultaneously achieves little accuracy loss or no accuracy loss on both datasets. Then, we
conduct experiments on WIDER FACE and COCO, making it possible to detect 80 classes of COCO as well as face with a single model, without labeling faces on COCO. We observe no accuracy loss on COCO and WIDER FACE.

We believe the flexible and general framework provides a solid approach for cross-dataset training and continual learning for academic research and industrial applications.

\section{Related Work}
\noindent
\textbf{Object Detection}
In the past decade, object detection has been involving rapidly. Deep CNNs (Convolutional Neural Networks) bring object detection to a totally new era. The two-stage R-CNN series~\cite{girshick2014rich, girshick2015fast, ren2015faster, lin2017feature, he2017mask} and single-stage detectors like SSD~\cite{liu2016ssd}, YOLO~\cite{redmon2016you} and RetinaNet~\cite{lin2018focal} are among the most popular frameworks. We choose RetinaNet as the detection framework, because it achieves state-of-the-art performance with focal loss to balance the positive and negative samples.
RetinaNet is also robust and effective on detection, instance segmentation and keypoint tasks.  Backbones are also important for detection performance. ResNets~\cite{he2016deep} are widely used in state-of-the-art object detectors. For mobile-friendly applications,  MobileNets~\cite{howard2017mobilenets} and its variants~\cite{zhang2017shufflenet,sandler2018mobilenetv2} are widely adopted for benchmarking. Object scale is another important research topic in object detection. Various detection frameworks adopt various solutions for scale~\cite{liu2016ssd,fu2017dssd,redmon2017yolo9000, lin2017feature}. Feature Pyramid
Network~\cite{lin2017feature} is an efficient multi-scale representation learning method for object detection. It achieves consistent improvement on a number of detection frameworks. RetinaNet with ResNets and FPN is the current leading framework on benchmarks like COCO.

\noindent
\textbf{Cross-dataset Training}
In terms of cross-dataset training, the most related work is Recurrent Assistance~\cite{perrett2017recurrent},  where cross-dataset training is used for frame-based action recognition
during pre-training stage.  
Since the number and identity of classes are different in each dataset, the authors propose to generate a new dataset by label concatenation, 
where labels from different datasets are simply concatenated to form a hybrid dataset with more labels.  
They have found that improved accuracy and faster convergence can be achieved by pre-training on similar datasets using label concatenation.
A similar label concatenation is also adopted in our paper, but in the training stage we must consider the duplication and conflict of different datasets instead of a straight-forward concatenation.

Another related work is integrated face analytics networks~\cite{li2017integrated}, where multiple datasets annotated for different tasks (facial landmark, facial emotion and face parsing) are used to train an integrated face analysis model, avoiding the need of building a fully labeled common dataset for all the tasks. The performance is boosted by explicitly modelling the interaction of different tasks. It belongs to the scope of multi-task learning. Although we have a similar motivation for cross-dataset training, we are the first to apply the general idea of cross-dataset training to object detection tasks, where different problems need to be addressed compared with those previous works.

\noindent
\textbf{Multi-task Learning}
Another related research topic is multi-task learning~\cite{argyriou2007multi, long2017learning}. Among the seminal works, the most related one is MultiNet~\cite{teichmann2018multinet}, where classification, detection and semantic segmentation are jointly trained on a single dataset. But it is quite different from our scenario, where multiple datasets are jointly trained for similar task but different object classes. In the multi-task learning community, there is still no appropriate algorithms for cross-dataset training.

\section{Cross-dataset Training}

\begin{figure*}
    \centering 
    \includegraphics[width=0.98\linewidth]{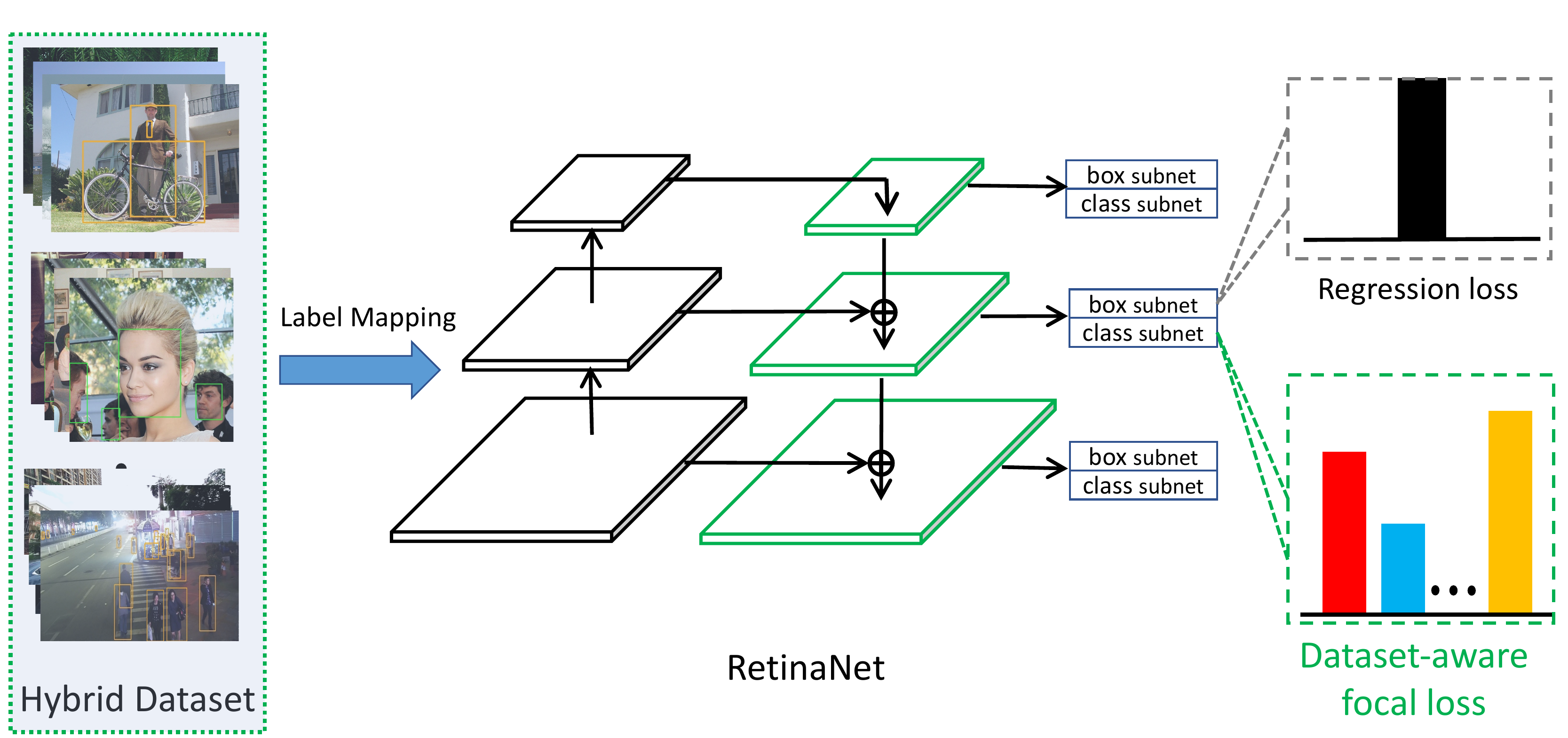} 
    \caption{Overall structure of the proposed cross-dataset training method. Existing datasets are merged into a hybrid dataset. Then RetinaNet is adopted as the detector. A shared regression loss is adopted over all classes after box subnet, while a dataset-aware focal loss is used to enable the training on the hybrid dataset after the class subnet. Different colors in the dataset-aware focal loss imply different classes from merged dataset. }
    \label{fig:cross}
\end{figure*}

\subsection{Detection Baseline}

Cross-dataset training for object detection aims to detect the union of all the classes across different existing datasets without additional  labelling efforts. Considering duplicate labels and possible conflicts across datasets, simply concatenating labels of all datasets is unreasonable and  may  cause degraded performance. Using RetinaNet as baseline, we propose a novel cross-dataset training scheme specially designed for object detection. The key components include label mapping and dataset-aware classification loss, which is a revised version of focal loss in RetinaNet. The overall structure of the proposed cross-dataset training method is shown in Figure~\ref{fig:cross}.

RetinaNet is a widely used one-stage object detector, where focal loss is proposed to solve foreground-background class imbalance during training. RetinaNet contains two parts: the backbone network and two subnetworks for object classification and bounding box regression. As in the original paper, we adopt the Feature Pyramid Network (FPN)~\cite{lin2017feature} to facilitate multi-scale
detection and the details of pyramid configuration are exactly the same as RetinaNet. Specifically, we use five pyramid levels of 256 channels, where the three levels with larger spatial sizes are calculated from the corresponding stage of ResNet-50~\cite{he2016deep} and MobilenetV2 ~\cite{sandler2018mobilenetv2}. We choose these two backbones to demonstrate that the proposed cross-dataset training scheme can be applied to detectors with networks of various sizes. In Section~\ref{sec:experiment}, we observe similar results on both large and small backbones.  

\subsection{Label Mapping}

\begin{figure}
    \centering 
    \includegraphics[height=6cm, width=7.5cm]{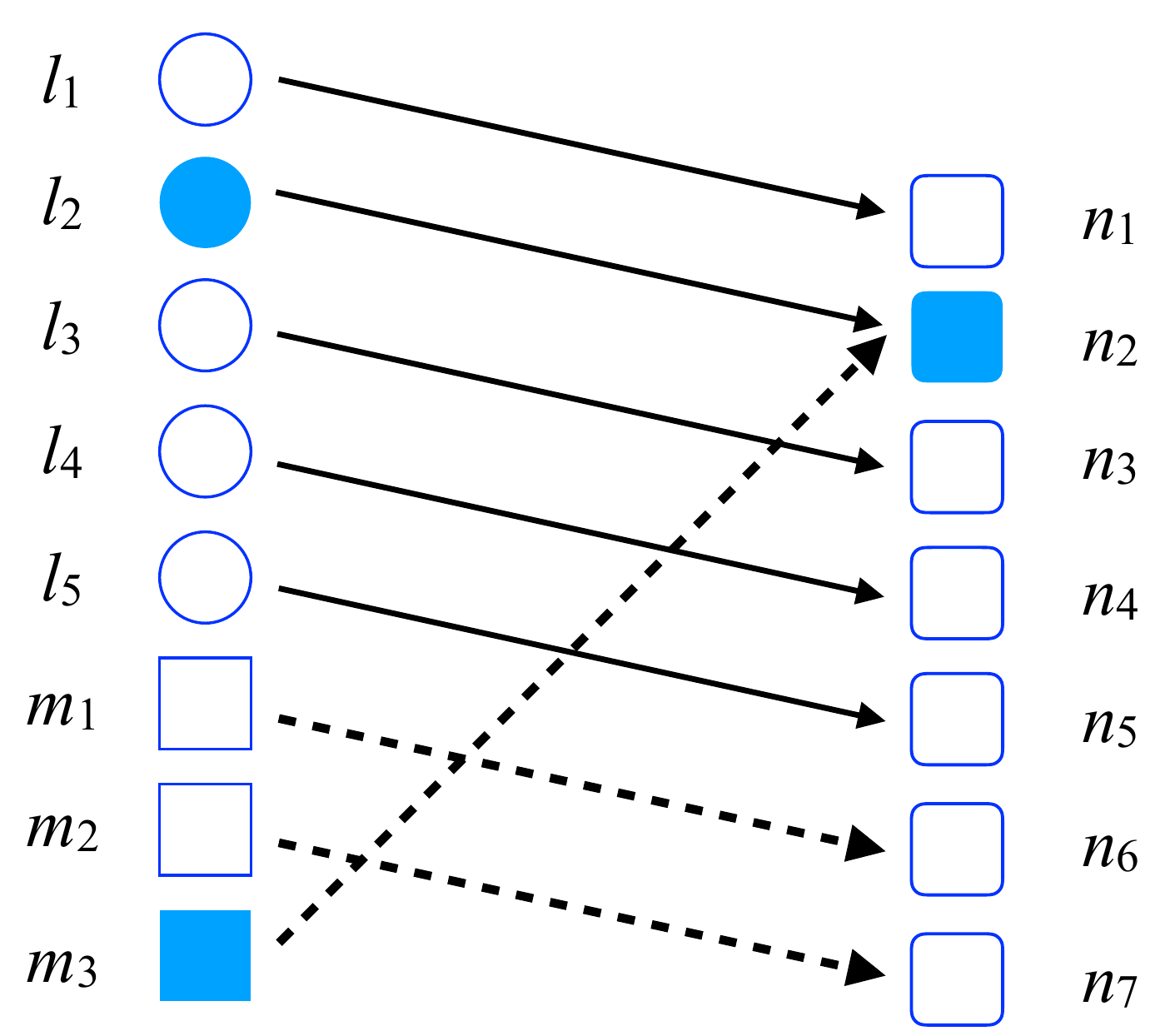} 
    \caption{A simple example of label mapping. Two datasets with labels $l_{1},l_{2},l_{3},l_{4},l_{5}$ (labels are represented by circles) and $m_{1},m_{2},m_{3}$ (represented by rectangles) are merged into a new hybrid dataset with labels $n_{1}, n_{2}....n_{7}$ (represented by round rectangles). The solid shapes in blue indicate that the labels belong to the same class that can be merged into one class in label mapping operation. Other labels stay unchanged during mapping.}
    \label{fig:label_mapping}
\end{figure}

As mentioned above,  duplicated or semantically consistent labels across datasets need to be merged. Then a hybrid dataset can be generated through label concatenation, where the source label of each image in origin dataset is kept to target dataset for using dataset-aware focal loss.  These two steps can be summarized as a label mapping process. All the old labels are mapped to a new set of labels where only unique labels are kept. A simple example of label mapping is given in Figure~\ref{fig:label_mapping}. Assuming we have two datasets, whose labels are $l_{1}, l_{2}, l_{3}, l_{4}, l_{5}$ and $m_{1}, m_{2}, m_{3}$ respectively. Label $l_{1}$ and $m_{3}$ have the same or similar meaning, so they will be mapped to the same new label $n_{2}$ in the hybrid dataset. By this label mapping procedure, we can obtain a new hybrid dataset where labels are consistent without duplicated ones.

\subsection{Dataset-aware focal loss}

The loss function for cross-dataset object detection needs to be carefully designed because of the possible conflicts of positive and negative samples. For example, negative samples from face detection dataset may be positive samples for human bodies, which means face detection dataset is a conflicting dataset for human detection. To accommodate this problem, we propose a new type of focal loss which is dataset-aware. The original focal loss for binary classification is 

\begin{align}\label{key}
FL(p_{t}) &= - \alpha (1 - p_{t})^{r} \log(p_{t})  \\
p_{t} &= \left\{
\begin{aligned}
p &, \text{if } y=1 \\
1-p &, \text{otherwise}
\end{aligned}
\right.
\end{align}

In the above $y$ specifies the ground truth class label and $p$ is the estimated probability for the label $y=1$. Other notations follow the RetinaNet paper. The proposed dataset-aware focal loss is 

\begin{align}\label{key}
FL(p_{t}) &= - \alpha (1 - p_{t})^{r} \log(p_{t})  \\
p_{t} &= \left\{
\begin{aligned}
p &, \text{if } y=1 \\
1 &, \text{if } y\not=1 \text{ and from a conflicting dataset} \\
1-p &, \text{otherwise}
\end{aligned}
\right.
\end{align}

\begin{figure}[t]
    \centering 
    \begin{subfigure}{0.98\linewidth}
    \includegraphics[width=0.98\linewidth]{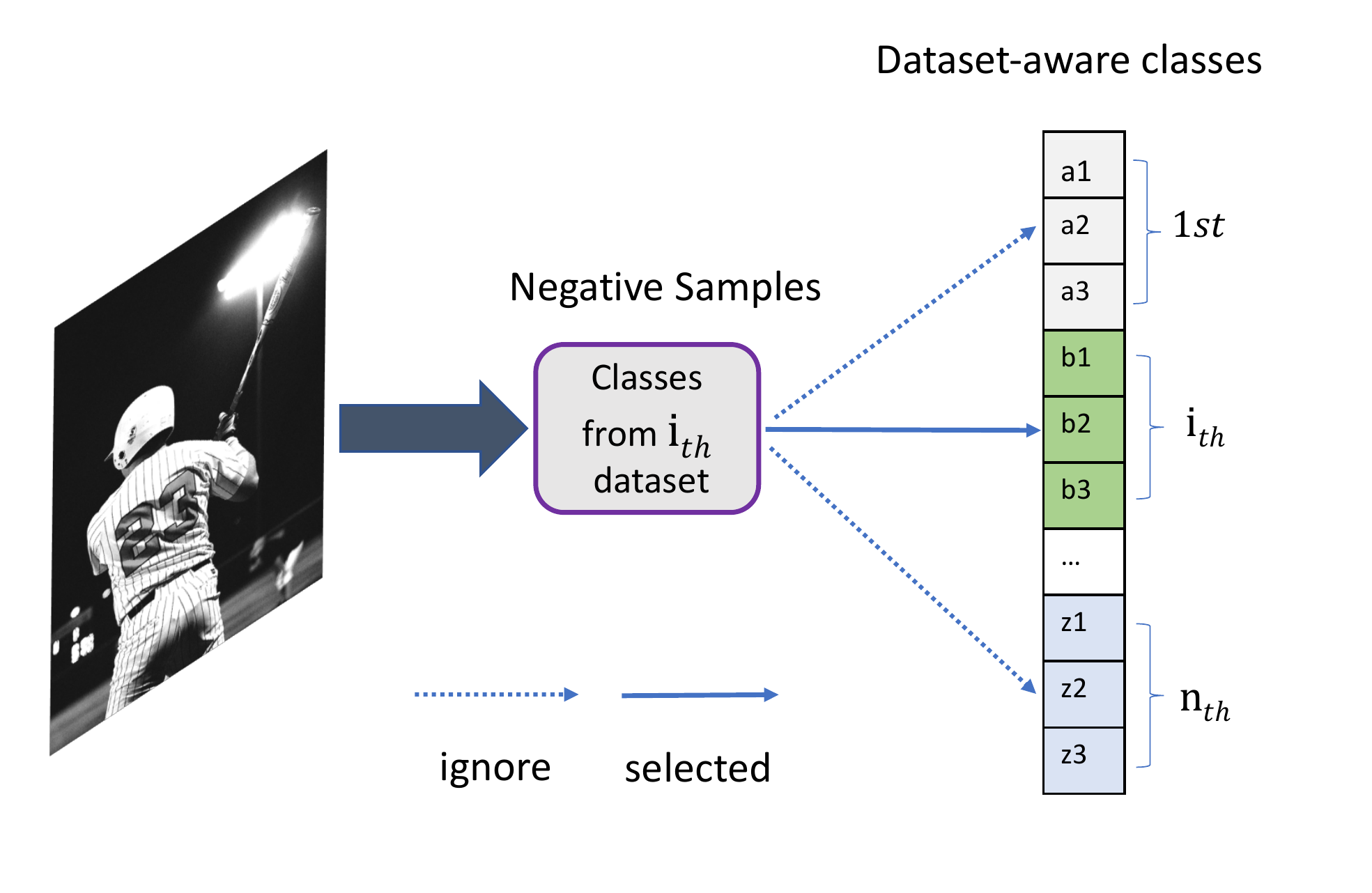} 
    \caption{Overview of dataset-aware focal loss.\label{fig:focal_loss_a}}
    \end{subfigure}
    \begin{subfigure}{0.98\linewidth}
    \includegraphics[width=0.98\linewidth]{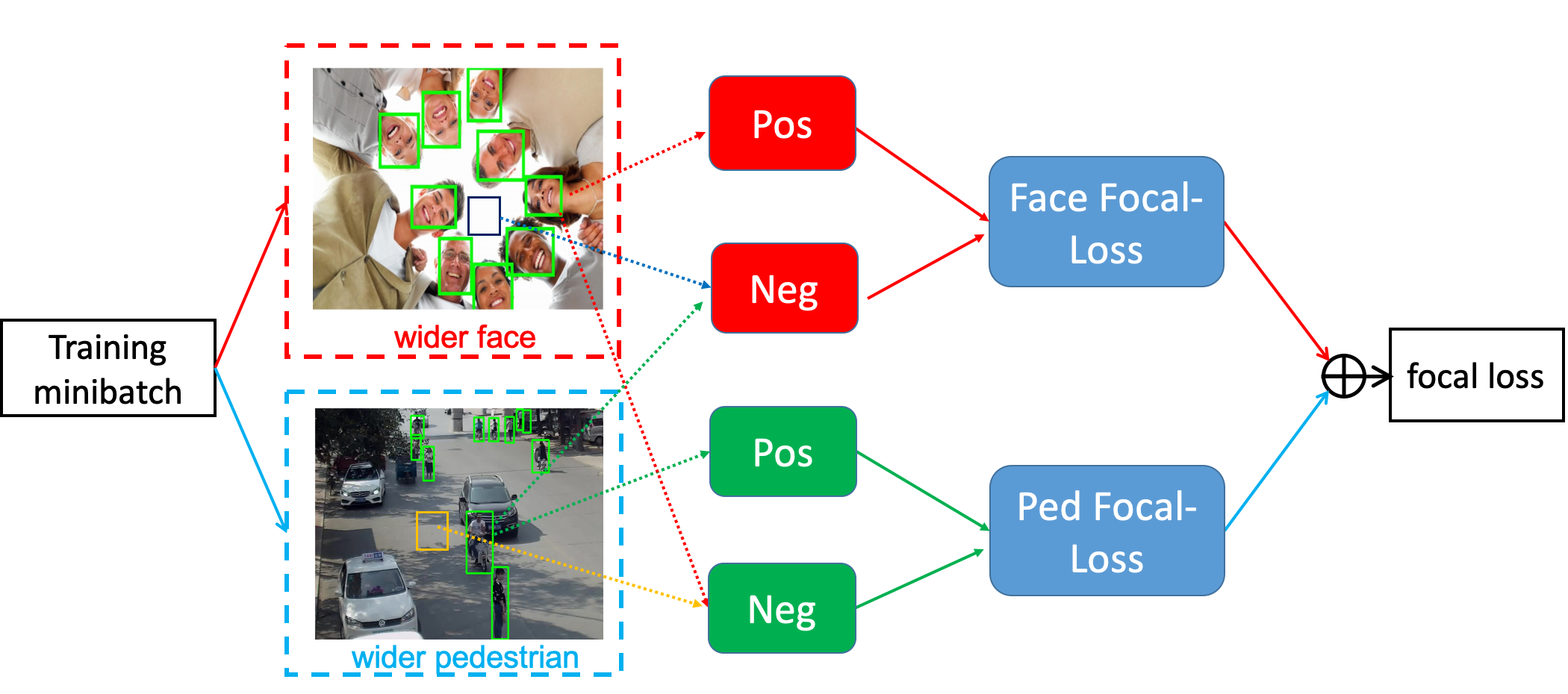} 
    \caption{An example of dataset-aware focal loss for face and pedestrian datasets.\label{fig:focal_loss_b}}
    \end{subfigure}
    \caption{(a) Overview of dataset-aware focal loss. The original dataset information is kept for all the samples in the hybrid dataset. Negative Samples from $i_{th}$ dataset will only contribute to the focal loss of object classes from $i_{th}$ dataset. (b) An illustration of dataset-aware focal loss for face and pedestrian datasets. These two datasets are considered to be conflicting because negative samples in one dataset may include objects being labeled as ground-truth in the other dataset. In this case, positive samples are generated according to the new labels after label mapping and can be processed according to the standard focal loss, but negative samples are not shared when calculating the focal loss of object classes from different datasets. Meanwhile, ground-truth patches from one dataset are included to unshared negative examples of other datasets.} 
    \label{fig:focal_loss}
\end{figure}

The most important modification of dataset-aware focal loss is that loss values corresponding to negative samples from conflicting datasets are set to zero. When training across multiple datasets, negative samples from one dataset will only contribute to the focal loss of those object classes within exactly the same dataset as shown in Figure~\ref{fig:focal_loss_a}. A simple illustration of dataset-aware focal loss is given in Figure~\ref{fig:focal_loss_b}. There exist conflicts between face and pedestrian datasets because their negative samples may include objects with labels from the other dataset, which causes confusion if using normal loss functions for classification. In
dataset-aware focal loss, negative samples are not shared across different datasets. So loss values of negative samples from face dataset are set to zero when calculating focal loss for the class pedestrian. Positive samples from different datasets are generated together according to their own ground truth labels, so there exist no conflicts and their loss values are calculated using the standard focal loss.
Through this dataset-aware approach, we can avoid the confusion caused by conflicting datasets. Futhermore, positive examples from one dataset are regarded as negative ones in other dataset to enrich the negative information for cross-dataset training.

In the Section \ref{sec:experiment}, we will show dataset-aware classification loss are necessary for stable training and good performance.

\section{Experiments}
\label{sec:experiment}
We conduct experiments to validate the feasibility of cross-dataset training with different detection tasks, and also provide detailed ablation study in this section.

\subsection{Experimental Setting}
\subsubsection{Datasets}In our experiments, three distinct detection tasks, i.e. face detection, general object detection and pedestrian detection are taken into consideration. Each task is associated with one or two different datasets.

\begin{table*}[t]
\centering
\footnotesize \setlength{\tabcolsep}{9.5pt}
\begin{threeparttable}
\begin{tabular}{c|c|c|c|ccc|ccc}
\toprule[1.5pt]
Data &Backbone &ratios & scales &AP &AP$_{50}$ &AP$_{75}$ &AP$_{\it S}$ &AP$_{\it M}$ &AP$_{\it L}$\\
\hline

COCO &ResNet-50 &[0.5, 1, 2] &[4,5.03,6.35]&35.5 &54.9 &38.0 &19.8 &39.2 &46.2 \\
COCO &ResNet-50 &[0.5, 1, 2]&[2, 3, 4]&35.4 &56.4 &37.3 &19.8 &38.7 &45.6 \\
COCO &MobileNetV2 &[0.5, 1, 2]&[4, 5.03, 6.35]&30.0 &48.6 &31.4 &16.2 &31.9 &39.3 \\
COCO &MobileNetV2 &[0.5, 1, 2]&[2, 3, 4] &29.5 &48.0 &30.6 &16.1 &31.3 &39.3 \\
COCO+FACE &ResNet-50 &[0.5, 1, 2] &[2, 3, 4]&35.2 &56.1 &37.1 &20.9 &38.3 &45.4 \\
COCO+FACE &MobileNetV2&[0.5, 1, 2]&[2, 3, 4]&29.5 &49.0 &30.4 &16.2 &31.4 &39.5 \\
COCO+PED &ResNet-50 &[0.5, 1, 2] &[2, 3, 4]&35.2 &55.6 &36.9 &20.2 &38.0 &44.9 \\
COCO+FACE+PED &ResNet-50 &[0.5, 1, 2] &[2, 3, 4]&35.0 &56.0 &36.7 &20.8 &38.2 &44.6 \\
\bottomrule[1.5pt]
\end{tabular}
\end{threeparttable}
\caption{The single and cross-dataset training COCO results on minival subset. All models are trained and tested with 800 pixel (shorter edge) input. FACE indicates WIDER FACE dataset, while PED indicates WIDER Pedestrian dataset.}
\label{tab:table1}
\end{table*}

\begin{table*}[t]
\centering
\footnotesize \setlength{\tabcolsep}{9.5pt}
\begin{threeparttable}
\begin{tabular}{c|c|c|c|ccc|ccc}
\toprule[1.5pt]
Data &Backbone &ratios & scales &AP &AP$_{50}$ &AP$_{75}$ &Easy &Medium &Hard \\
\hline

PED &ResNet-50 &[1.25,2.44] &[2, 3]&48.37 &84.26 &50.24 &- &- &- \\
PED &MobileNetV2 &[1.25,2.44] &[2, 3]&43.73 &81.94 &41.1 &- &- &- \\
FACE &ResNet-50 &[1.25,2.44] &[2, 3]&- &- &- &95.21 &93.7 &87.88 \\
FACE &MobileNetV2 &[1.25,2.44] &[2, 3] &- &- &- &93.0 &90.98 &84.64 \\
FACE+PED &ResNet-50 &[1.25, 2.44] &[2, 3]&48.32 &84.26 &49.71 &94.58 &93.22 &87.59 \\
FACE+PED &MobileNetV2&[1.25, 2.44]&[2, 3]&43.86 &81.0 &42.1 &92.24 &90.41 &84.63 \\
COCO+PED &ResNet-50 &[0.5, 1, 2] &[2, 3, 4]&52.46 &85.64 &56.6 &- &- &-\\
COCO+FACE+PED &ResNet-50 &[0.5, 1, 2]&[2, 3, 4]&51.44 &85.03 &54.9 &95.02 &93.76 &86.66\\
\bottomrule[1.5pt]
\end{tabular}
\end{threeparttable}
\caption{The PED and WIDER FACE results on val subset.}
\label{tab:table2}
\end{table*}

The task of face detection aims to detect faces in images. We use the popular WIDER FACE~\cite{yang2016wider} dataset for experiments. It consists of 32,203 images and 393,703 faces with large variability in scale, pose and occlusion. The dataset is split into training ($40\%$), validation ($10\%$) and test ($50\%$) set. Besides, the images are divided into three levels (easy, medium and hard subsets) according to the difficulty of detection. The images and annotations of training and validation set are available online, and all the results of face detection in our experiments are reported on the validation set.

To verify our method on general object detection task, we conduct a series of experiments on the COCO ~\cite{lin2014microsoft} detection dataset. For training, we use standard \texttt{coco-2017-train}, which contains 115k images. The evaluation results are reported on the 5k val images(\texttt{coco-2014-minival}). We also carry out many experiments on PASCAL VOC dataset to prove the effectiveness of our method when the number of classes increases a lot. 

For pedestrian detection task, the WIDER Pedestrian~\cite{wider-pedestrian} dataset with 11,500 training images, 5,000 validation images and 3,500 testing images is used in our experiments. We report our results on the 5,000 validation images in our experiments for thorough comparison.

\subsubsection{Implementation Details}
We choose RetinaNet as our detection baselines. ResNet-50 and MobileNetV2 are used as our backbones. Images are resized such that their scale (shorter edge) is 800 pixels, the same as original paper. Our models are trained using SGD with 0.9 momentum, 0.0001 weight decay and batch size 32. The first two epochs are set as warm-up for stable training. The maximum number of epochs is 20 and we use 0.04 learning rate for the first 10 epochs, and then continue train for another two 5 epochs with learning rate of 0.004 and 0.0004.

Intuitively, the size of objects may vary a lot during cross-dataset training procedure, so the default setting of anchor scales during training may no longer suitable for cross-dataset scenarios. For instance, when combining with face or pedestrian detection tasks, the default anchor scales of \{4, 5.03, 6.35\} in COCO setting are too large to detect small faces or pedestrians, but the larger anchor scales are needed to detect large scale objects in COCO and PASCAL VOC evaluation. So adjustments on training scales and ratios are necessary for cross-dataset training to ensure the performance. The final anchor scales are the product of the simplified anchor scales we proposed and strides in each pyramid level. As new anchor scales are introduced in our cross-dataset training experiments, we choose default settings and newly-revised ones together as our baselines for a fair comparison.

According to the reason above, for COCO, we use anchors at three aspect ratios of \{1:2, 1:1, 2:1\} and the scales are set as \{2, 3, 4\} and {\{4, 5.03, 6.35\}}. And for PASCAL VOC, ratios of \{1:2, 1:1, 2:1\} and scales of \{4, 5.03, 6.35\} are selected as our baseline setting.

Similarly, we have three different baselines in our experiments for face detection tasks. The corresponding combinations of ratios and scales are \{ \{1.25\}; \{2, 3\}\}, \{ \{1.25, 2.44\}; \{2, 3\}\} and \{ \{0.5, 1, 2\}; \{2, 3, 4\} \}
 
For pedestrian detection, the ratio and scales are set to \{2.44\} and \{2, 3\}, respectively. Finally, we use the ratios of \{1.25, 2.44\} and scales of \{2, 3\} in FACE-PED cross-dataset training and adopt the ratios of \{0.5, 1, 2 \} and scales of \{2, 3, 4\} in PED-COCO and FACE-COCO-PED cross-dataset training.

\subsection{Results and Comparison}
We compare the performance of the proposed cross-dataset training with our baseline results. We consider several different cross-dataset training settings including face with COCO, face with pedestrian, pedestrian with COCO and all mixed datasets. The same network structure and same training strategy are used for fair comparison. As for evaluation metrics, we evaluate the COCO-style Average Precision (AP) for COCO general object detection task and pedestrian detection task. For the WIDER FACE dataset, the validation set has easy, medium and hard subsets, which roughly correspond to large, medium and small faces respectively. 

\subsubsection{Baselines}

\textbf{WIDER FACE: } We use RetinaNet to train face detectors as our baselines.
Different from original configuration, we set serveral different settings for better performance considering the relatively fixed ratio of the face. Figure \ref {fig:val_roc} and Table \ref {tab:table3} shows our results.

\textbf{COCO: } We first reproduce the original COCO RetinaNet results using PyTorch. However, the default anchor scales are not suitable for face detection. So we re-configure the anchor scales to get adaptive results both in COCO and WIDER FACE. For fair comparison, we retrain the
COCO results with new anchor scales. Table \ref{tab:table1} shows our baseline results. Our baseline result (35.4 AP) is very close to the result (35.7 AP) of the original paper.

\textbf{PASCAL VOC: } PASCAL VOC dataset are applied to carry out cross-dataset training with COCO dataset to verify our idea on the general object detection task. The anchor ratios are same with COCO settings. Table \ref {tab:table5} shows the results of our baselines.

\textbf{WIDER Pedestrian: } We first follow the same anchor aspect ratio of \{2.44\} as used in ~\cite{zhang2017citypersons} for accurate pedestrian detection by applying RetinaNet framework. For a more fair comparison with cross-dataset training, we also add serveral anchor settings same with cross-dataset training to act as baselines as well. 

\subsubsection{Cross-dataset training}

\begin{figure*}[!t]
  \centering
  \begin{subfigure}[b]{.3\textwidth}
    \includegraphics[width=\textwidth]{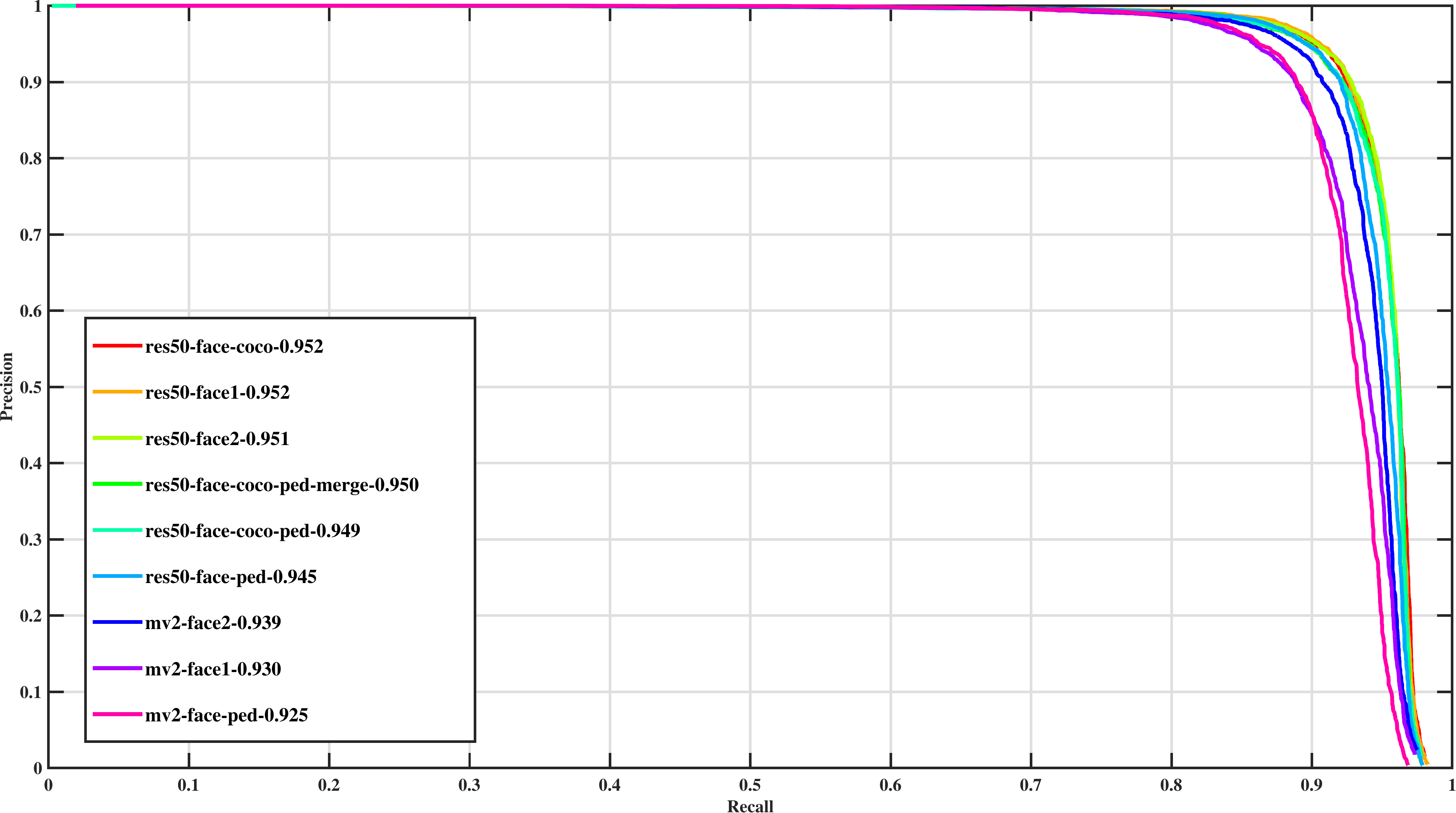}
    \caption{Easy.}
    \label{fig:val_easy}
  \end{subfigure}
  \begin{subfigure}[b]{.3\textwidth}
    \includegraphics[width=\textwidth]{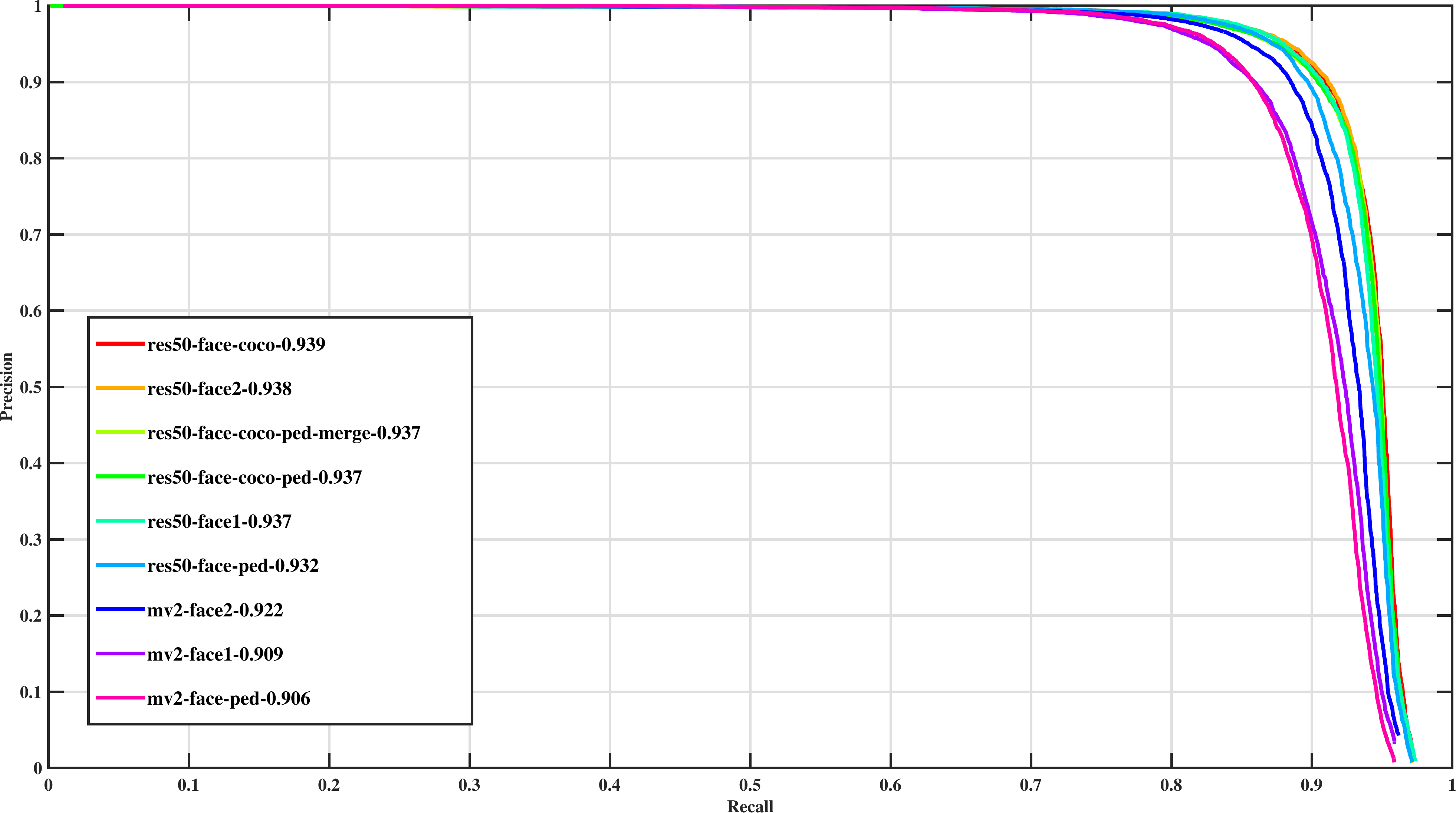}
    \caption{Medium.}
    \label{fig:val_medium}
  \end{subfigure}
  \begin{subfigure}[b]{.3\textwidth}
    \includegraphics[width=\textwidth]{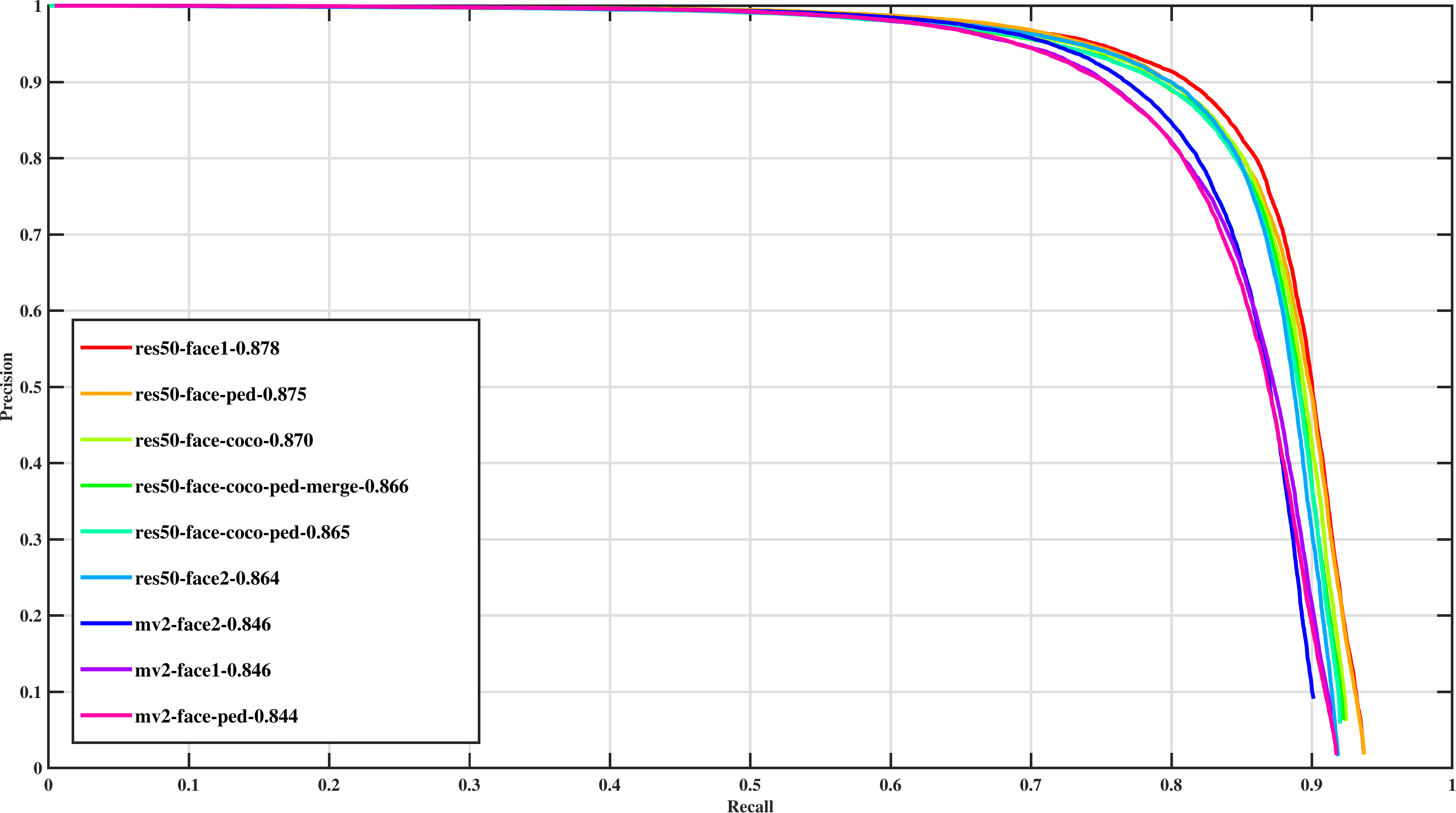}
\caption{Hard.}
    \label{fig:val_hard}
  \end{subfigure}
  \caption{Precision-recall curves on WIDER FACE val set. Results for baselines and cross-dataset training of two different backbones, ResNet50(res50) and MobileNetV2(mv2) are drawn. Legends with "face1" indicate baseline settings for WIDER FACE with WIDER Pedestrian, while those with "face2" are for WIDER FACE with COCO.}
  \label{fig:val_roc}
\end{figure*}

\textbf{WIDER FACE with WIDER Pedestrian: }
In these experiments, WIDER FACE and WIDER Pedestrian are mixed for cross-dataset training. From Table \ref {tab:table2}, it gets the 48.32 AP, almost the same as the baseline result (48.37 AP) in the pedestrian detection task. For the face detection task, similar results are shown in Figure \ref{fig:val_roc}. The results clearly demonstrate that cross-dataset training strategy is powerful in face and pedestrian detection tasks without degrading in performance. 

\textbf{WIDER FACE with COCO: }
In these experiments, we mix WIDER FACE with COCO for cross-dataset training. As shown in Table \ref {tab:table1}, it gets an AP of 35.2 using ResNet-50 backbone. Compared to the baseline results, cross-dataset training results only drop little in COCO minival set. As reported in Figure \ref {fig:val_roc}, it maintains performance on three subsets. For MobileNetV2 backbone, similar results are shown in Table \ref {tab:table1}, where the AP of COCO minival is 29.5 vs 29.5 for cross-dataset settings and baseline.

\textbf{WIDER Pedestrian with COCO:}
Results for adding pedestrian to COCO are included in Table \ref {tab:table2}. Note that the person class in COCO are similar with pedestrian, we need merge the class. It gets 35.2 AP similar to the original result. Moreover, it gets higher performance in pedestrian detection task because COCO dataset also contains pedestrians, which brings stronger pedestrian feature. 

\begin{table}[t]
\centering
\small
\footnotesize \setlength{\tabcolsep}{5.5pt}
\begin{threeparttable}
\begin{tabular}{c|c|c|c|ccc}
\toprule[1.5pt]
Data & ratios &scales &AP &Easy &Medium &Hard \\
\hline
FACE  & [1.25] &[2,3] &- &94.7 &93.27 &87.60 \\
FACE  &[0.5,1,2]&[2,3,4]&- &95.14 &93.80 &86.48 \\
FACE  &[1.25,2.44] &[2,3]&- &95.21 &93.7 &87.88 \\
COCO  &[0.5,1,2] &[4,5,6]&35.5 &- &- &- \\
COCO  &[0.5,1,2] &[2,3,4] &35.4 &- &- &- \\
PED   &[2.44] &[2,3,4] &50.82 &- &- &- \\
PED   &[1.25,2.44] &[2,3] &48.37 &- &- &- \\
PED   &[0.5,1,2] &[2,3,4] &50.0 &- &- &- \\
F+C   &[0.5,1,2]&[2,3,4] &35.2 &95.23 &93.97 & 87.06 \\
F+P   &[1.25,2.44] &[2,3] &48.32 &94.58 &93.22 &87.59 \\
\bottomrule[1.5pt]
\end{tabular}
\end{threeparttable}
\caption{Different anchor settings results with backbone of ResNet-50.}
\label{tab:table3}
\end{table}

\textbf{WIDER Pedestrian, COCO and WIDER FACE: } We add WIDER Pedestrian and WIDER FACE to COCO dataset for cross-dataset training. We can conclude that adding more datasets for cross-dataset training don't harm the performance in each single dataset. Observation from Table \ref {tab:table2} suggests that the AP rises from 50.0 to 51.44 for PED and slightly declines for WIDER FACE. Table \ref {tab:table1} shows that the AP for COCO suffers minor drop from 35.4 to 35.0, which can be explained by the fact that mixed part of two pedestrian categories may bring disturbation for COCO detection.

\textbf {COCO with PASCAL VOC: } Finally, we add PASCAL VOC to COCO dataset to validate the effectiveness of our training method when more classes are influenced or merged in cross-dataset training pipeline. We can observe from Table \ref {tab:table5} that COCO dataset maintains the performance, while PASCAL VOC dataset get even better performance on cross-dataset scenarios,  benefitting from more data introduced by COCO dataset.

The experiments above demonstrate the effectiveness of cross-dataset training in general object detection tasks. By cross-dataset training, we can utilize the existing datasets to detect more classes without extra works of labeling all the classes in different datasets.

\begin{table}[t]
\centering
\small
\footnotesize \setlength{\tabcolsep}{9.5pt}
\begin{threeparttable}
\begin{tabular}{c|c|c|c|c}
\toprule[1.5pt]
Data & Backbone &Merge &COCO &PED \\
\hline
C\_P   & R50      &no   &35.2 & 51.0 \\
C\_P   & R50     &yes &35.3 & 51.43 \\
C\_P   & Mv2      &no   &29.1 & 46.75 \\
C\_P   & Mv2     &yes &29.1 & 47.21 \\
F\_C\_P & R50   &no &35.1 &50.99  \\
F\_C\_P & R50   &yes &35.0 &51.45  \\

\bottomrule[1.5pt]
\end{tabular}
\end{threeparttable}
\caption{Merge labels results. C\_P indicate COCO and PED datasets. F\_C\_P indicates FACE, COCO and PED datasets.}
\label{tab:table4}
\end{table}

\begin{table}[t]
\centering
\small
\footnotesize \setlength{\tabcolsep}{9.5pt}
\begin{threeparttable}
\begin{tabular}{c|c|c|c}
\toprule[1.5pt]
Data & Backbone &COCO &VOC \\
\hline
COCO  & R50 &35.5 & - \\
COCO  & Mv2 &29.5 & -\\
VOC   & R50 &- & 83.0\\
VOC   & Mv2 &- & 76.4\\
COCO+VOC & R50 &35.4 &87.5 \\
COCO+VOC  & Mv2 &29.5 &83.1\\

\bottomrule[1.5pt]
\end{tabular}
\end{threeparttable}
\caption{The COCO and VOC results in val set. }
\label{tab:table5}
\end{table}

\subsubsection{Ablation Study}
\textbf{Merge semantically identical classes is necessary for cross-dataset training?} 
In our cross-dataset training settings, those classes with identical semantic meanings, such as person and pedestrian, are combined during training. We conduct different experiments of whether to merge similar classes to verify the validity of that strategy. From Table \ref {tab:table4}, our PED results can get an improvement of about 0.5 AP in cross-dataset train after merging classes via semantic information. Actually, if we don't merge those similar categories, they are mutually exclusive and the definition of negative examples may be confused, doing harm to our classification task training.

\textbf{How the anchor scales influence the results for cross-dataset training ?} 
In this experiment, we have explored the effects of different anchor scales on cross-dataset training. Each detection task has specific anchor scales for better performance. Compared to face and pedestrian detection tasks that require smaller anchor scales, COCO general object detection task needs larger anchor scales to detect various objects. For a more fair comparison of experimental results, we keep anchor scales and ratios same on the single dataset and cross-dataset training. Table \ref {tab:table4} shows the cross-dataset training maintains the performance when we use same anchor settings with baseline.

\begin{table}[t]
\centering
\small
\footnotesize \setlength{\tabcolsep}{8.5pt}
\begin{threeparttable}
\begin{tabular}{c|c|c|ccc}
\toprule[1.pt]
Data & Backbone &AP &Easy &Medium &Hard \\
\hline
FACE  &Res50 &- &95.21 &93.7 &87.88 \\
FACE  &Mv2 &- &93.00 &90.98 &84.64 \\
PED   &Res50 &48.37 &- &- &- \\
PED   &Mv2 &43.73 &- &- &- \\
F+P   &Res50 &48.32 &94.58 &93.22 &87.59 \\
F+P   &Mv2 &43.86 &92.5 &90.6 &84.41 \\
\bottomrule[1.pt]
\end{tabular}
\end{threeparttable}
\caption{Different backbone results of face and pedestrian detection tasks. F , P indicate FACE and Pedestrian datasets. All the anchor scales in these experiments are \{2, 3\}. Mv2 is short for MobileNetV2.}.
\label{tab:table6}
\end{table}

\begin{figure*}
\centering
\includegraphics[width=0.98\textwidth]{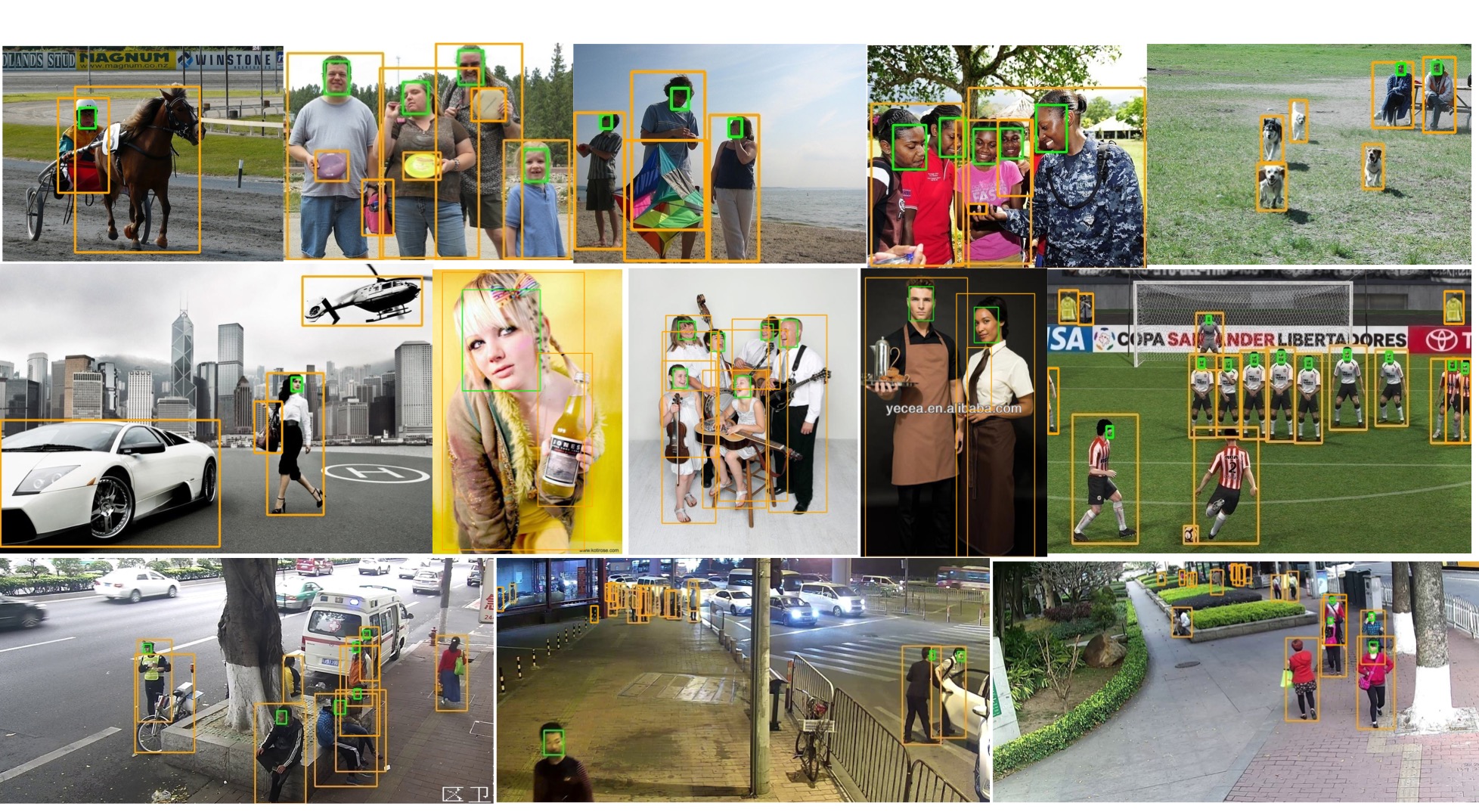}
\caption{More results of our cross-dataset training on different datasets. The three rows are samples selected from COCO, WIDER-Face and Pedestrian Dataset, respectively.}
\label{fig:example}
\end{figure*}

\textbf{How important are the backbones for cross-dataset training ?}
We conduct experiments on two different backbones, i.e. ResNet-50 and MobileNetV2, to illustrate that our cross-dataset training method works not because of the redundant parameters but a general effective pipeline for multi-dataset training. Results in Table \ref {tab:table6} support our assumption. Cross-dataset training can maintain performance for both two datasets and similar results are given when we choose smaller backbones using cross-dataset training. We can see from Table \ref {tab:table6} that for both backbones with large (ResNet-50) and limited (MobileNetV2) parameters, the proposed cross-dataset training setting does not undermine the performance on both datasets, indicating that our training pipeline is a universal solution, not depending on the redundancy of the network.

\section{Applications}
Cross-dataset training is a general concept. We demonstrate how it can be used to train a single unified object detector with multiple datasets. When data with new class labels arrives, existing data does not need to be labeled again.

This method can be extended to other applications as well. As long as the datasets have different classes or the same class but different domains, cross-dataset training can be generalized to train a single model. For example, 360 degree (in-plane rotation, i.e.\ \emph{roll}) support is a difficult problem in face detection. Only using data augmentation can not solve this problem well. But with cross-dataset training, we can add orientation supervision during data augmentation. Different orientations can be viewed as different classes. The trained model can detect faces in 360 degrees and predict the face orientations.

\section{Conclusion}
In this paper, we introduce cross-dataset training for object detection. It aims to jointly train multiple datasets labeled with different object classes. We propose a training scheme using label mapping and cross-dataset classification loss. Experiments on various popular datasets and backbones prove the effectiveness of our approach. With cross-dataset training, we make it possible to detect the class union of multiple datasets with a single model without accuracy loss. 
We expect this general training method to be used in three scenarios: 1) object detection research that utilizes existing object detection datasets, 2) industrial applications that are usually faced with increasing classes, 3) life long learning where data with new class labels continuously arrive.

{\small
\bibliographystyle{ieee_fullname}
\bibliography{detection}
}

\end{document}